\documentclass{article}
\usepackage{ijcai22}

\usepackage{times}
\usepackage{soul}
\usepackage{url}
\usepackage[hidelinks]{hyperref}
\usepackage[utf8]{inputenc}
\usepackage[small]{caption}
\usepackage{graphicx}
\usepackage{amsmath}
\usepackage{amsthm}
\usepackage{xcolor}
\usepackage{booktabs}
\usepackage{algorithm}
\usepackage{hyperref}
\usepackage{multirow}
\usepackage{subfigure}
\usepackage{algorithm}
\usepackage{algpseudocode}

\newcommand{\citet}[1]{\citeauthor{#1} \shortcite{#1}} \newcommand{\citep}{\cite} 

\title{Causal Intervention-based Prompt Debiasing for Event Argument
Extraction}

\author{
Jiaju Lin$^1$
\and
Jie Zhou$^{2}$\and 
Qin Chen$^1$  \\
\affiliations
$^1$School of Computer Science and Technology, East China Normal University\\
$^2$School of Computer Science, Fudan University
\emails
jiaju\_lin@stu.ecnu.edu.cn,
jie\_zhou@fudan.edu.cn,
qchen@cs.ecnu.edu.cn,
}

\begin{document}
\maketitle
\begin{abstract}
Prompt-based methods have become increasingly popular among information extraction tasks, especially in low-data scenarios. By formatting a finetune task into a pre-training objective, prompt-based methods resolve the data scarce problem effectively. However, seldom do previous research investigate the discrepancy among different prompt formulating strategies. In this work, we compare two kinds of prompts, name-based prompt and ontology-base prompt, and reveal how ontology-base prompt methods exceed its counterpart in zero-shot event argument extraction (EAE) . Furthermore, we analyse the potential risk in ontology-base prompts via a causal view and propose a debias method by causal intervention. Experiments on two benchmarks demonstrate that modified by our debias method, the baseline model becomes both more effective and robust, with significant improvement in the resistance to adversarial attacks.
\end{abstract}

\section{Introduction}
Event argument extraction (EAE) plays an independent role in natural language processing. It has become widely deployed in downstream tasks like natural language understanding and decision making\cite{zhang2022enhancing}. 
During the past few years, a pretrain-finetuning paradigm of large language models has achieved great success. However, it is unavoidable to train a new task-specific head for every new appearing type with plenty of labeled instances, that leads to `data hungry' and sets barriers to real-word implementation. 

Fortunately, a novel paradigm, namely `prompt', provides a promising approach to address the data scarce problem \cite{liu2021pre}. By formatting a finetune task into a pre-training objective, prompt-based methods have become the best-performed baselines especially in low-data scenarios. Nonetheless, how to design a proper prompt is still an open problem. 
Although automatic prompts generating methods have experiences a great surge in the last years, manual prompts still dominate information extraction area. Current state-of-the-art prompt-based methods \cite{li-etal-2021-document} are mainly based on manual designed prompts. Besides, previous works \cite{ma-etal-2022-prompt} also verify the ineffectiveness of auto-prompts in EAE. Based on these observations, we wonder \textit{ which is the better prompt design strategy and how this prompt facilitates extraction.}

In this paper, to answer the above questions, we divide current manual prompts into two categories: 1) \textbf{name-based prompts} which are formed by concatenating names of the arguments belonging to an event type. 2) \textbf{ontology-based prompts} which are derived from the event ontology, the description of an event type in natural language. We carry out quantitative analysis on predictions of these two prompts and find that, compared with name-based prompts, ontology-based prompts can provide additional syntactic structure information to facilitate extraction. By filtering the improper potential arguments, ontology-based prompts improve model's overall performance. Nevertheless, every coin has its two sides. The hidden risks are introduced along with the beneficial information. We theoretically identify the spurious correlation caused by ontology-based prompts from a causal view. Based on the structural causal model, we find that the model trained with ontology-based prompts may have bias on entities that share the common syntactic role with the argument name in the prompt, e.g. both the entity in the sentence and argument name in the prompt are subject. 

We further propose to conduct causal intervention on the state-of-the-art method. Via backdoor adjustments, intervened model rectifies the confounder bias stem from the similar syntax. Experiments are performed on two well-know benchmarks, namely RAMS and WikiEvents. The enhancements in performance demonstrate the effectiveness of our proposed approach. Moreover, we evaluate the robustness of our method via exposing the model to adversarial attacks and noise in training. The results show that modified by our adjustments, the model becomes more robust than ever before.

Our contributions are threefold:
\begin{itemize}
    \item We propose a causal intervention-based prompt debiasing model for event argument extraction based on bias found by investigating how ontology-based prompts work in zero-shot event argument extraction task.
    \item We rethink the prompt-based event argument extraction in a causal view to analyze the causal effect among different elements and reduce the biases via backdoor adjustments. 
    \item Extensive experiments on the cutting-edge method and datasets demonstrate the effectiveness and robustness of our method.
\end{itemize}

\section{Pilot Experiments and Analysis}
To investigate the work mechanism of prompts for extraction, we investigate two varieties of the most advanced prompt based-method PAIE \cite{ma-etal-2022-PAIE} in zero-shot event argument extraction. In this setting, events are split into two parts, top $n$  most common `seen events' and  the rest `unseen events'. Both training and validation sets are composed of `seen events' only, while the test set merely contains the `unseen events'. In our experiment, we set $n$ =4. 

\subsection{Model}
We apply the most cutting-edge method PAIE \cite{ma-etal-2022-PAIE}, an efficient prompt-based extraction model for both sentence-level and documental-level EAE. It introduces span selectors for extraction rather than forms it as a generation task. Following PAIE's original setting, we investigate two different prompt creation strategies:
\begin{itemize}
    \item[1] \textbf{name-based prompt creation}, which concatenates names of the arguments belonging to an event type.
    \item[2] \textbf{ontology-based prompt creation}, where prompts are derived from the event ontology, which depicts the interactions among prompts in natural language.
\end{itemize}

\subsection{Datasets}
We conduct experiments on three common datasets in event argument extraction: 
RAMS \cite{ebner-etal-2020-multi}, WikiEvents\cite{li-etal-2021-document} and ACE05\cite{doddington2004automatic}.
The former two datasets are annotated in documental-level. RAMS includes 139 event types and 65 semantic role. In the WikiEvents dataset, 50 event types and 59 semantic roles are defined. Although it renders the coreference links of arguments, we only consider conventional arguments. While ACE05 is a multilingual dataset providing sentence-level event annotation. Here we use its English set for our experiments. For data split of ACE05, we follow the preprocessing procedure of DyGIE++ \cite{wadden2019entity}.


\begin{table}[th]
\begin{tabular}{c|cll|cll|cll}
\hline
               & \multicolumn{3}{c|}{ACE}  & \multicolumn{3}{c|}{RAMS} & \multicolumn{3}{c}{WikiEvents} \\ \hline
name based     & \multicolumn{3}{c|}{37.3} & \multicolumn{3}{c|}{23.4} & \multicolumn{3}{c}{26.3}       \\
ontology based & \multicolumn{3}{c|}{43.5} & \multicolumn{3}{c|}{34.3} & \multicolumn{3}{c}{32.0}       \\ \hline
\end{tabular}
\caption{Overall results for pilot experiments. The metric used here is span-level F1.}
\label{tab:conVsFull}
\end{table}

\begin{figure}
    \centering
     \subfigure[Spurious Error Ratio]{
        \begin{minipage}{7cm}
         \centering
         \includegraphics[width=\textwidth]{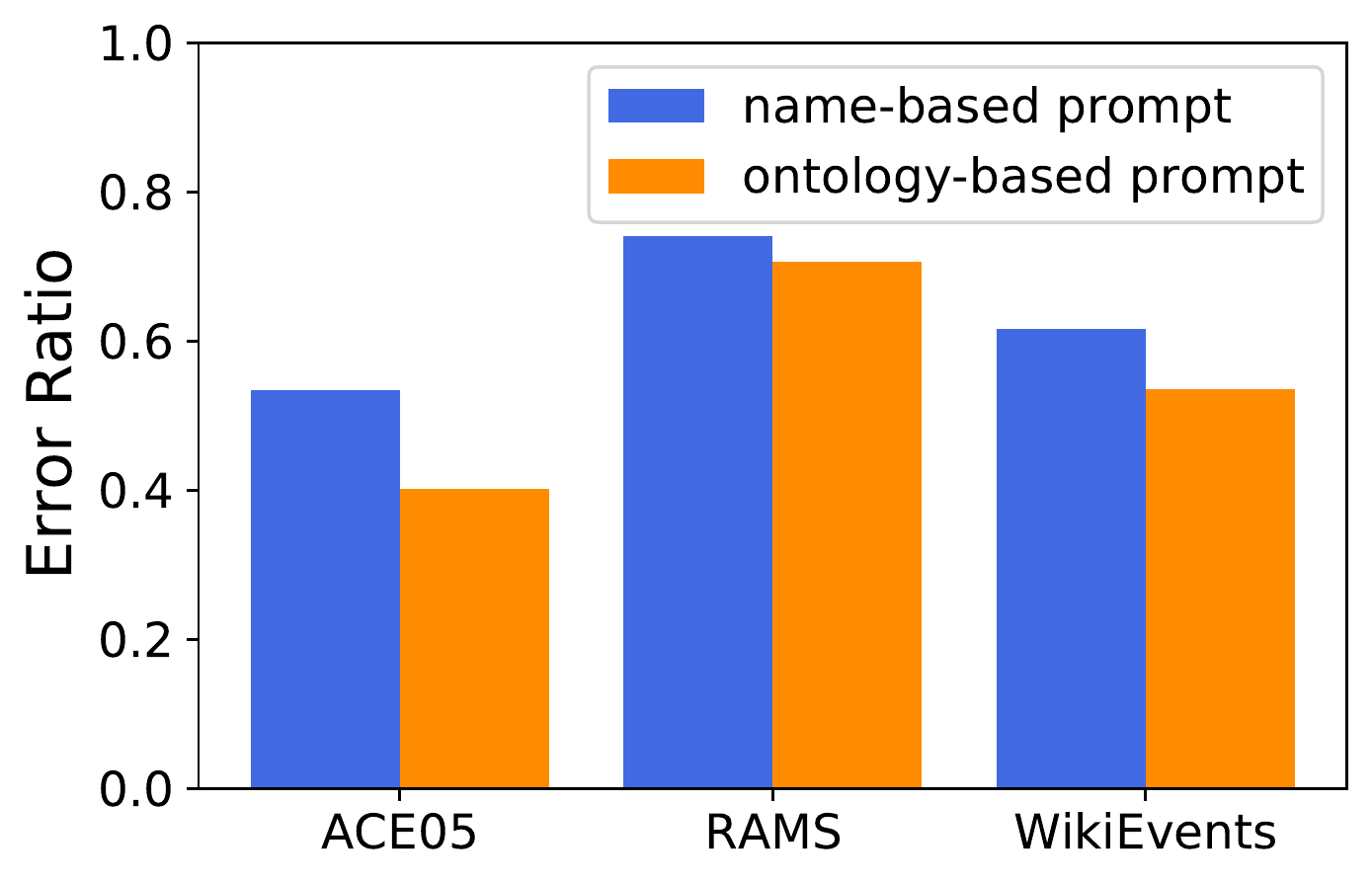}
        \end{minipage}
         \label{fig:SpuriousError}
     }
    \hfill
    \subfigure[Syntactic Role Matching Ratio]{
        \begin{minipage}{7cm}
         \centering
         \includegraphics[width=\textwidth]{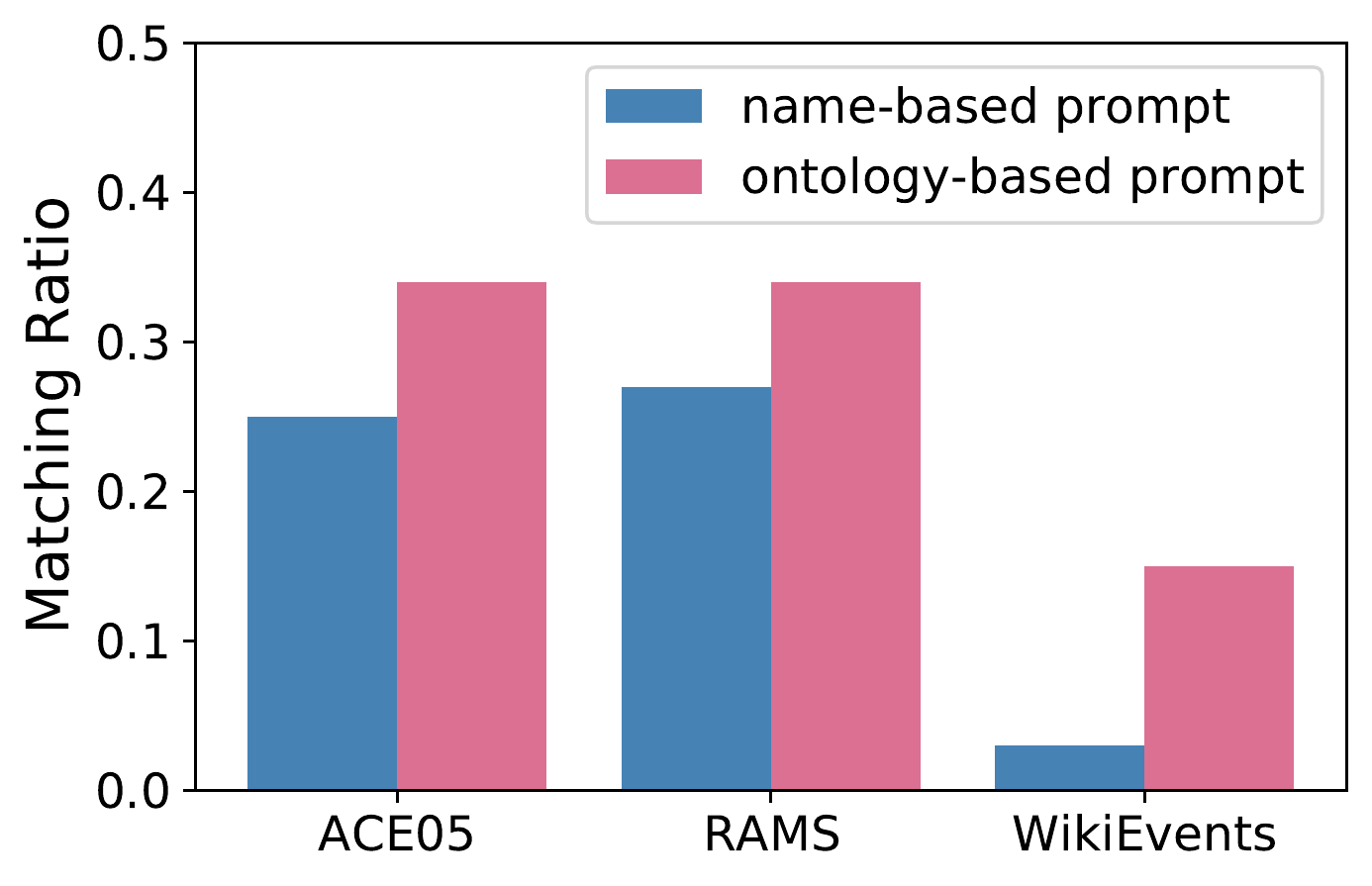}
        \end{minipage}
        \label{fig:CompositionMatch}
     }
    \caption{Detailed comparison between name-based prompt and ontology-based prompt method.}
    \label{fig:pilotAnalysis}
\end{figure}

\begin{table*}[th]
\centering
\begin{tabular}{c|cc}
\hline
sample 1 \label{sample1}      & \multicolumn{2}{c}{... {\color{blue} Trump} appeared to $\rm  <t> urge </t> $a {\color{red} U.S. adversary}
suspected of criminal activity ...}                                                                                                                             \\ \hline
            & \multicolumn{1}{c|}{name-based}                                                                                       & ontology-based                                                                                                   \\ \hline
prompts     & \multicolumn{1}{c|}{\underline{recipient} \underline{communicator} \underline{ place}}                                                                     & \begin{tabular}[c]{@{}c@{}}  \underline{communicator} communicated remotely\\ with \underline{recipient} about topic at \underline{place}\end{tabular} \\
predictions & \multicolumn{1}{c|}{\begin{tabular}[c]{@{}c@{}}communicator:{\color{red} U.S. adversary}\\ recipient: { \color{red} U.S. adversary}\end{tabular}} & \begin{tabular}[c]{@{}c@{}}communicator: {\color{blue} Trump}\\ recipient :{ \color{red} U.S. adversary}\end{tabular}                         \\ \hline
sample 2       & \multicolumn{2}{c}{$\rm ... They  <t> hired </t>$ {\color{purple} Heede} to a baseline greenhouse gas inventory ...}                                                                                                                                                 \\ \hline
            & \multicolumn{1}{c|}{name-based}                                                                                       & ontology-based                                                                                                   \\ \hline
prompts     & \multicolumn{1}{c|}{\underline{employee} \underline{placeofemployment} \underline{place} }                      & \begin{tabular}[c]{@{}c@{}}\underline{employee} started working at\\ \underline{placeofemployment} in \underline{place}\end{tabular}                 \\ 
predictions & \multicolumn{1}{c|}{employee: {\color{purple} Heede} }                                                                                   & employee: They                                                                                                   \\ \hline
\end{tabular}
\caption{Case study on the difference between two prompts.}

\label{tab:pilotCaseStudy}
\end{table*}

\subsection{Results and Analysis}
As shown in Table \ref{tab:conVsFull}, on each dataset, ontology-based method exceeds name-based remarkably by at least 5.8 point in F1 score. To further investigate how ontology-based prompts success, we conduct a quantitative error analysis on each prediction set. 
Imitating previous work \cite{das-etal-2022-automatic}, we define the mistakes when the ground truth is `None' but the model extracts some arguments as spurious role error, for it reflects that the model extracts these none-existing arguments dependent on some spurious relations.
Figure \ref{fig:SpuriousError} shows that on each  dataset, ontology-base prompts generate less spurious role errors. Furthermore, by analysing the syntactic role of predicted arguments, we observe that syntactic similarity between the instance sentence and ontology-based prompt promotes the extraction. Specifically, for every prediction, we compare the dependency label, like `nsubj' or `dobj', of each predicted argument in the original sentence and its argument name in the ontology by spaCy\footnote{https://github.com/explosion/spaCy}. The matching degree is shown in Figure \ref{fig:CompositionMatch}. On every dataset, the dependency labels of predictions from ontology-based prompt match the event ontology better. And there is always a gap around 6 point in the matching ratio. 

With above observations, it is reasonable to conclude that compared with name-based prompt, ontology-based prompts impose more syntactic level constraint when extracting, rectifying some improper predictions by their syntactic roles. As case study displayed in Table \ref{tab:pilotCaseStudy} sample 1, `Trump' is the subject of the input  sentence, matching the role of `communicator' in the ontology-based prompt. Depending on this clue, the model extracts the right communicator `Trump'. Although ontology-based prompt partly solves the long-standing problem about entity type relying, it introduces new vulnerability. In Table \ref{tab:pilotCaseStudy} sample 2, owing to the word `They' is in the subject position, the same as where `employee' locates in the ontology prompt, the model mistakenly identifies `They' as 'employee'. We will discuss such kind of potential risks from a causal view in the next chapter. 


\section{Background}
\subsection{Causal Inference}
Causal inference is a technique that can identify undesirable biases and fairness concerns in benchmarks \cite{kusner2017counterfactual,vig2020investigating,feder2021causal}. Causal inference describes the causal relations between variables via Structural Causal Model (SCM), then identify confounders and spurious relations for bias analysis. Eventually, true causal effects can be estimated by eliminating biases with causal intervention.

\textbf{SCM} The structural causal model \cite{pearl2000models} is a directed acyclic graph, defining a set of causal assumptions on the variables of interest in a problem. It describes dependencies between variables with a graphical model $G=\{V,f\}$, where each node represents a variable, and every edge represents causal effects $f$ between two variables.

\textbf{Causal Intervention} To estimate the true causal effects between a pair of variables $(X,Y)$, we introduce causal intervention, which fixes $X=x$ and removes all arrows towards the node $X$, denoted as $do(X=x)$. By such adjustment, $P(Y|do(X=x))$ represents an unbiased, de facto causal effects of treatments $X$ on outcome $Y$ \cite{pearl2009causal}.

\textbf{Backdoor Criterion} To adjust the influence from confounder, backdoor criterion is on of the most important tool for causal intervention. Given a fork structure in an SCM like $X \leftarrow Z \rightarrow Y$
, to estimate the real causal effects from $X=x$ on $Y$, we stratify the confounder $Z$ and calculate the effects by:
\begin{equation}
\nonumber
\begin{split}
        P(Y|do(X=x&))= \sum_{z}P(Y|X=x,Z)P(Z)
\end{split}
\end{equation}
where $P(Z)$ can be estimated from data or given in advance. In this way, the relation between $Z$ and $X$ is cut off, resulting in $X$ independent of $Z$.

\subsection{Prompt-based Event Argument Extraction} \label{PAIE}
Current Prompt-based Event Argument Extraction methods can be formulated as follows, given an input sentence of $|x|$ tokens $x=\{x_1,x_2,...,x_{|x|}\}$, a labelled trigger span $(t_{s},...,t_{e})$, and a prompt sequence $pt$ with $|pt|$ tokens, $pt=\{pt_1,pt_2...,pt_{|pt|}\}$, the target is to identify the relevant argument spans then classify them into predefined arguments types. Mainstream methods feed $x$ and $pr$ into a encoder-decoder model and utilize the output representation of $pt$ to conduct extraction. 

As mentioned before, we apply the state-of-the-art model PAIE, as the backbone of our method. According to PAIE's architecture, the encoder only receive input sentence, the fusion of the sentence and the prompt happens in decoding stage.
\begin{equation}
\begin{split}
    H_{X}^{enc}&=\text{Encoder}(x) \\ 
    H_X  &= \text{Decoder}( H_{X}^{enc};H_{X}^{enc}) \\
    H_{pt} &= \text{Decoder}( pt; H_{X}^{enc} ) 
\end{split}
\end{equation}

Then a role-specific span selector $\{\phi_{start},\phi_{end}\}$ is trained to extract argument spans from $x$.
\begin{equation}
\begin{split}
        \phi_{start} &= W^{start}\phi_a \\
        \phi_{end}   &= W^{end}\phi_a
\end{split}
\end{equation}
where $ \{W^{start},W^{end}\} \in R^h $ are trainable parameters, and $\phi_a$ is the hidden state of argument name $a$ in $H_{pt}$.

Eventually, the spans are selected from $H_X$:
\begin{equation}
    \begin{split}
        p^{start}&=\text{Softmax}(\phi^{start}H_X) \\
        p^{end}&=\text{Softmax}(\phi^{end}H_X)
    \end{split}
\end{equation}
and the loss function is defined as:
\begin{equation}
    \mathcal{L} = \sum_{k} -(\log p^{start}_k s_k+ \log p^{end}_k e_k)
\end{equation}
Here $k$ represents the index of the argument span, $s_k,e_k$ means the start and end position of a span respectively.

\begin{figure}[t!]
    \centering
    \includegraphics[width=0.9\linewidth]{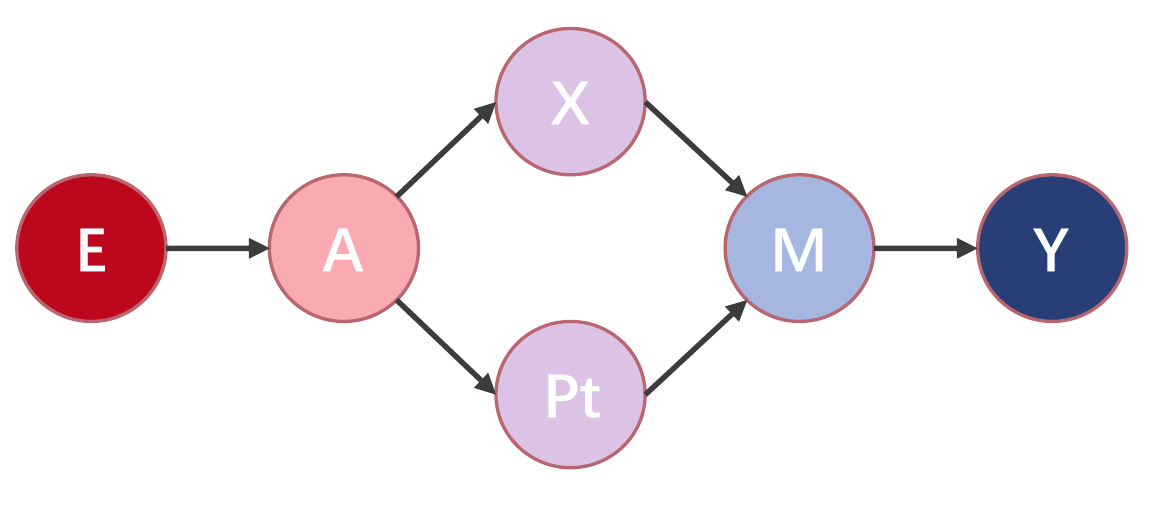}
    \caption{Structual casual model for prompt-based event argument extraction.}
    \label{fig:SCM}
\end{figure}

\section{Rethinking in a Causal View}
\subsection{Structural Causal Model For Prompt-Based Event Argument Extraction}
Figure \ref{fig:SCM} shows the structural causal model (SCM) for Prompt-Based Event Argument Extraction (PEAE). The main idea behind the SCM is that prompts generated from annotation guidelines naturally perform well in samples selected following the same guidelines. Specifically, there is a strong correlation between the syntactic structure of the labeled instance and that of the corresponding prompt. 
In the SCM, we describe PEAE with four variables: 1) an Event notion in human mind $E$ 2) an Event Ontology for Annotation $A$, which depicts the event by several arguments' interactions; 3) a Prompt $Pt$; 4) an Instance $X$, which is labelled following annotation ontology; 5) a Language Model $M$; 6) Extraction Results $Y$.
Given the variables mentioned above, a SCM can be formulated by following causal relations:
\begin{itemize}
    \item \textbf{Ontology Design.} The path $E \rightarrow A$ denotes how specialists design an event ontology $A$ by event notion $E$. The ontology are composed of a sew of arguments whose interactions consist an event and some verbs which can be deemed as examples of triggers.
    \item \textbf{Instances Annotation.} The path $A \rightarrow X$ describes the process when annotators labelled the sentences into instances $X$ with triggers and arguments, following the instructions of annotation ontology $A$.
    \item \textbf{Prompt Generation.} The path $A \rightarrow Pt$ represents the prompt generation procedure, where an annotation ontology $A$ is modified into a corresponding prompt $Pt$.
    \item \textbf{Prompt-based Extraction.} The collision structure $\{X,\ Pt\} \rightarrow M \rightarrow Y$ symbolizes the  argument extraction process where an instance $X$ and a prompt $Pt$ are fed into a language model $M$ as input, outputting an extraction result $Y$. 
\end{itemize}

\subsection{Spurious Relation in Prompt-Based Event Argument Extraction}
To build a robust EAE model, we need to estimate $P(Y|X,E)=P(Y|X,Pt)$, as describe by $\{X,\ Pt\} \rightarrow M \rightarrow Y$. Unfortunately, there exists a backdoor path $X \leftarrow A \rightarrow Pt $ between an instance  $X$ and language model $M$. That is to say, the learned effect of $X$ to $Y$ is confounded by $Pt$ (through $A$). 
For example, given an ontology \textit{\underline{`Manufacturer} manufactured or created or produced \underline{Artifact} using \underline{Instrument} at \underline{place}'} of event 'manufacture.artifact.n/a' and a trigger-labeled sentence \textit{`The \underline{company} also gives Saudis more flexibility and has \textbf{created} \underline{fast - track programs}...'}, the subject `company' will be annotated as 'Manufacture' and the object `fast-tract programs' of trigger `created' will be labeled as `Artifact'.
During training, if we directly copy the annotation ontology as the prompt, the model would mistakenly build a spurious correlation between the syntactic composition and labels. The 'company' is inferred as `Manufactureer' correctly for it plays a role of subject in the sentence, the same as `Manufactureer' played in the prompt. So is the `fast-tract programs'.
Consequently, the model can not learn the true causal relation $\{X,\ Pt\} \rightarrow M \rightarrow Y$, and lacks robustness.

\subsection{Causal Intervention to Resolve Spurious Correlation}
According to the backdoor criterion, we conduct the do-operation by intervening on the instance $X$:
\begin{equation}
\nonumber
     P(Y|do(X),E) 
          =\sum_{i}P(Y|x,E,Pt^{i})P(Pt^{i}|x,E)
\end{equation}
where $Pt$ is selected as the variable to block backdoor paths between $X$ and $M$.

\textbf{Estimating $P(Pt^{i}|x,E)$ via conditional generation.} 
The confounder distribution $P(Pt^i|x,E)$ is hard to calculate directly for $E$ is a hidden variable. Considering that $E$ can be induced to trigger-arguments' interaction and the trigger is labeled in $x$, we argue that $P(Pt^i|x,E) \propto G(Pt^i|t,Arg)$ which indicates a controllable generation task whose output text ought to cover the trigger $t$ in the instance and predefined arguments $Arg$ in the ontology. In our method, we apply a off-the-shelf keywords to sentence model MF \cite{wang-2021-mentionflag} to estimate $P(Pt^i|x,E)$. For each generated prompt sentence, we obtain $P(Pt^i|x,E)$ by a softmax calculation of the probability for each sentence.
\begin{equation}
    P(Pt^i|x,E) = \frac{exp(l_i)}{\Sigma_{j}exp(l_j)}
\end{equation}
Here $l_i$ represents the log likelihood for a generated prompt, calculated during the process of beam search generation.

\textbf{Learning De-biased Causal Relation.}
Once generating a prompt set $Pt_c=\{Pt_c^0,...,Pt_c^{|Pt_c|}\}$ and its distribution $P$, we merge them into prompts' representation. Considering the fact that in some cases generation model can not generate an output with proper semantic, we remain the original ontology-based prompt $Pt_o$ with a hyper-parameter $\lambda$:  
\begin{equation}
    \begin{aligned}
    H_p= &\lambda M(x;Pt_o)\ + \\
     &(1-\lambda) \sum_{i} P(Pt_c^i|x,E) M(x;Pt_c^i)
\end{aligned}
\end{equation}
Then we optimize a de-biased extraction model following the stpes mentioned in \ref{PAIE}. To be noticed, although in this paper, we only attempt causal intervention on PAIE model, our method can be applied on most prompt-based methods which calculate prompt's hidden state $H_{pt}$.

\begin{table*}[th]
\centering
\begin{tabular}{l|cc|ccl}
\hline
\multirow{2}{*}{Models}             & \multicolumn{2}{c|}{RAMS} & \multicolumn{3}{c}{WikiEvents} \\
        & Arg-I       & Arg-C       & Arg-I    & Arg-C    & Head-C   \\ \hline
EEQA \cite{du-cardie-2020-event}          & 46.4        & 44.0        & 54.3     & 53.2     & 56.9     \\
DocMRC \cite{liu-etal-2021-machine}        & -           & 45.7        & -        & 43.3     & -        \\
CUP \cite{lin2022cup}           &             & 46.5        &          & 59.9     &          \\ \hline
PAIE-name     & 53.2        & 47.5        & 62.4     & 61.7     & 65.7     \\
PAIE-ontology & 54.7        & 49.5        & 68.9     & 63.4     & 66.5     \\ \hline
PAIE-debias (Ours)   & \textbf{55.4}        & \textbf{50.2}        & \textbf{70.7}     & \textbf{65.5}     & \textbf{68.7}     \\ \hline
\end{tabular}
\caption{Overall performance in fully supervised setting.}
\label{tab:OverallPerformance}
\end{table*}

\section{Experiments}
\subsection{Experiment Setup}
\textbf{Datasets Selection.} Owing to the fact that the annotation guideline of ACE05\footnote{https://www.ldc.upenn.edu/sites/www.ldc.upenn.edu/files/english-events-guidelines-v5.4.3.pdf} is different from its ontology-based prompts, the causal effect $A \rightarrow Pt$ does not exist, which means the $X \rightarrow Y$ is not confound by $A$. Hence, we conduct de-bias experiments on two datasets, RAMS and WikiEvents. \\
\textbf{Baselines.} We compare our method with several state-of-the-art models:
\begin{itemize}
    \item Generation model: CUP \cite{lin2022cup} which introduces curriculum learning for documental argument extraction, where prompts are ontology-based.
    \item QA-based model: EEQA \cite{du-cardie-2020-event}, which is one of the first Question-Answering (QA) based model for event extraction and DocMRC \cite{liu-etal-2021-machine}, another QA-based model but enhanced by implicit knowledge transfer and explicit data augmentation. 
    \item Span-selection model: PAIE \cite{ma-etal-2022-prompt} the state-of-the-art method for EAE. Besides its sota version PAIE with ontology-based prompts (PAIE-ontology), we also investigate the performance of PAIE with name-based prompts (PAIE-name). 
\end{itemize}
For fairness of the comparison, in our experiments, all models are base-scaled, i.e., using a pretrained-language model as the backbone. \\
\textbf{Evaluation Metric} Two evaluation metrics are calculated to compare the performance of models. (1) Argument Idetification F1 score (Arg-I): an event argument is correctly identified if its offsets match any of the argument mentions. (2) Argument Classification F1 score (Arg-C): an event argument is correctly classified if its role type is also correct. Additionally, for WikiEvents dataset, we follow \cite{ma-etal-2022-prompt} and evaluate Argument Head F1 score (Head-C) which only concerns the matching of the head of an argument.
\textbf{Implementation Details.} We follow PAIE's implementation for a fair comparison. Specifically, we train each model on a single Nvidia GeForce RTX 3090. All parameters are optimized by AdamW \cite{loshchilov2017decoupled} algorithm. The optimal parameters are obtained based on the development set, then used for evaluation on test set. For hyperparameter $\lambda$ ,  we search it from 0 to 1 with a step of 0.1 .\\

\subsection{Fully Supervised Experiments}
Table \ref{tab:OverallPerformance} illustrates the comparison between our method and baselines. Our method performs best on both two datasets, with 0.7\% and 2.1\% gains in Arg-C respectively. The performance on Arg-I and Head-C has the same trend. Our method outperforms the best baseline PAIE-ontology with 0.7\% improvements in Arg-I score on RAMS and 1.8\% on WikiEvents. Moreover, we find that the span-selector model enhances overall performance by a large margin. Even with the sub-optimal prompt design, PAIE-name also performs better than the most advanced generation-based model, CUP, by at least 1 point in Arg-C. 

\begin{figure}
    \centering
     \subfigure[Zero-Shot Results for RAMS]{
        \begin{minipage}{5cm}
         \centering
         \includegraphics[width=\textwidth]{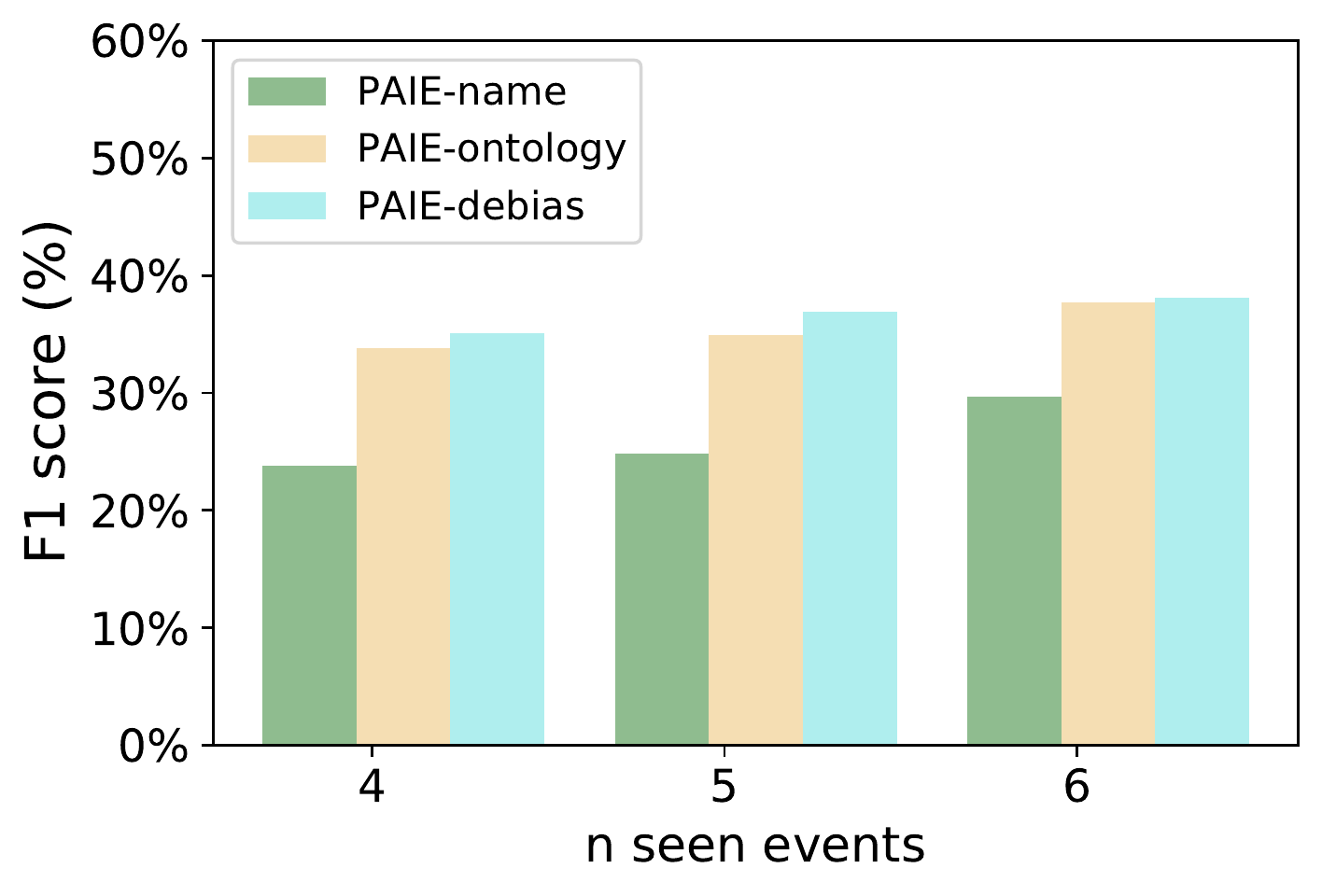}
        \end{minipage}
         \label{fig:transferRAMS}
     }
    \hfill
    \subfigure[Zero-Shot Results for WikiEvents]{
        \begin{minipage}{5cm}
         \centering
         \includegraphics[width=\textwidth]{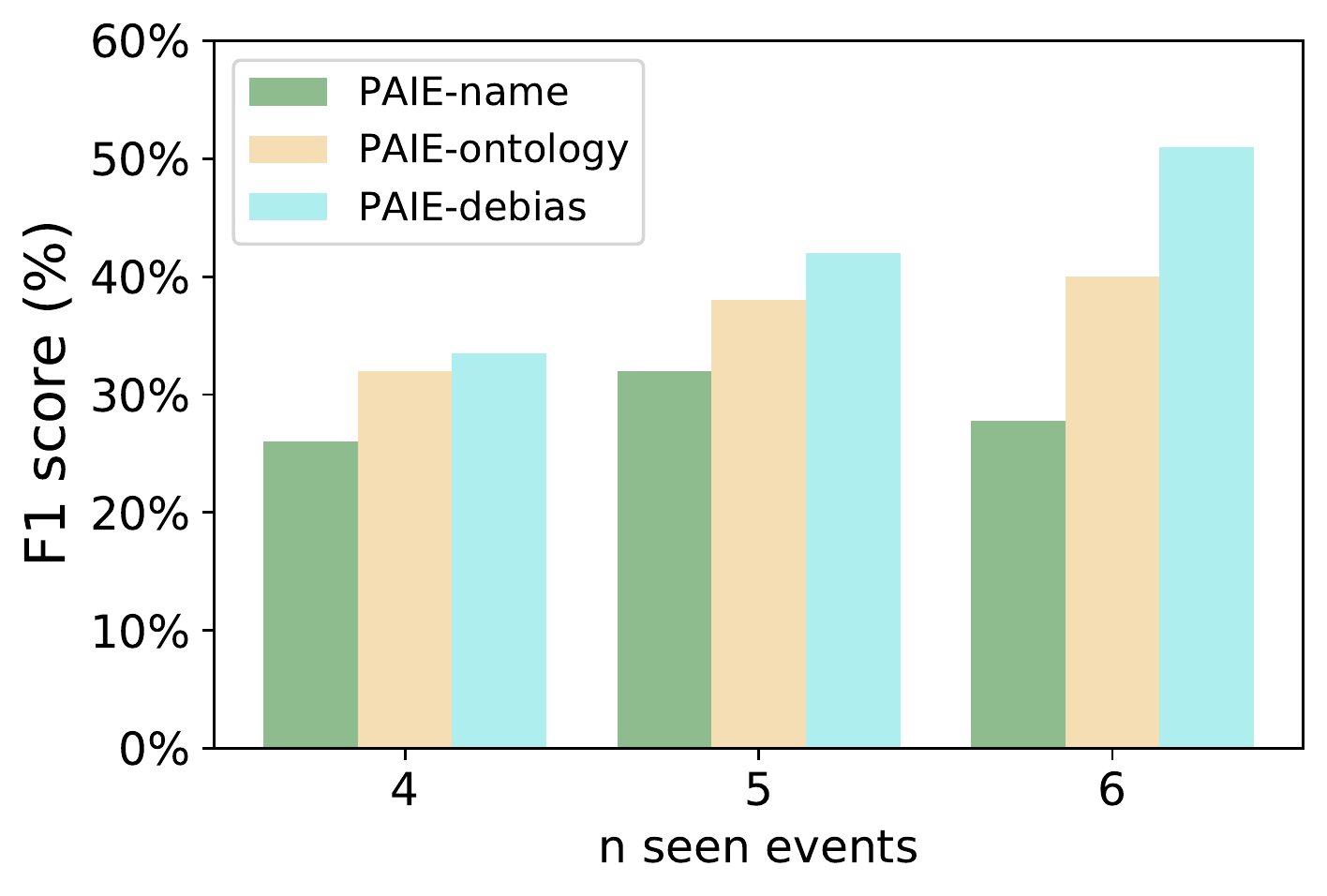}
        \end{minipage}
        \label{fig:transferWikiEvents}
     }
    \caption{Zero-shot results of two datasets.}
    \label{fig:zeroshotResults}
\end{figure}

\subsection{Zero-Shot Scenario}
We carry on experiments under zero-shot scenario with the same setting as in pilot experiments. To fully evaluate the zero-shot ability of our method, extensive experiments are conducted with $n$ ranging from 4 to 6. 
As shown in Figure \ref{fig:zeroshotResults}, our approach overtakes PAIE-ontology on both datasets under different seen-unseen divisions. Especially on WikiEvents, a sharp rise of performance happens with $n$ increases from 5 to 6, over 50\% in F1 on unseen types. 

\begin{table*}[t]
\centering
\begin{tabular}{l|ccc|ccc}
\hline
\multirow{2}{*}{RAMS} & \multicolumn{3}{c|}{PAIE-ontology} & \multicolumn{3}{c}{PAIE-debias} \\
                      & seen 4     & seen 5    & seen 6    & seen 4   & seen 5    & seen 6   \\ \hline
raw-prompt            & 34.4       & 34.9      & 37.7      & 35.5     & 36.7      & 38.1     \\
sub-optimal prompt    & 27.1       & 29.5      & 29.5      & 31.1     & 31.3      & 31.2     \\
$\Delta$             & -7.3      & -5.4      & -8.8      & -4.5($\uparrow$ 2.7)   & -5.4 (-)   & -6.9 ($\uparrow$ 1.9)  \\ \hline
\end{tabular}
\end{table*}
\subsection{Robustness to Prompts Variation}
One of the long-standing problems of manual-prompt based method is its sensitivity to prompts' design. Our method attempts to build a robust prompt-based model by generating a cluster of prompts and imitating their probability. To verify our model's resistance to prompts' change, we conduct robustness testing on RAMS and WikiEvents respectively. Regarding new prompt design, due to the fact that the syntactic structures of the automatic augmented prompts are still the same, we create new prompts manually. To generate the prompt with largest perturbation while the semantic meaning maintain, we follow these principles: 1) All argument names should be included in the new designed prompts. 2) The modification of the words used in new prompts should be as less as possible, 3) The syntactic discrepancy should be as much as possible, specifically, an argument's syntactic role in the new prompt should be different from the its old role. 4) For events where model performs poorly (<25\% in F1), which means  our modification may improve the performance, and whose event is hard to modify, we keep their prompts remained.


\section{Related Work}
\subsection{Event Argument Extraction}
Event argument extraction aims to extract argument spans from sentences with a given event type. Plenty of efforts have been devoted into this sphere since an early stage \cite{nguyen-etal-2016-joint-event,}. By converting documents into an unweighted graph, \citet{huang-jia-2021-exploring-sentence} use GAT to alleviate the role overlapping problem. \citet{huang-peng-2021-document}'s method capture the inter-dependencies among events to better model the event representation. Recently, there is a trend of formulating EAE into a question-answering problem \cite{du-cardie-2020-event,liu-etal-2020-event}. By inducing knowledge from pre-trained models, models can take advantage of large-scale language models better. \citet{wei-etal-2021-trigger} takes interactions among arguments into consideration and \citet{liu-etal-2021-machine}'s approach demonstrates the effectiveness of data augmentation. 

Based on the success of QA models, researchers explore prompt methods based on encoder-decoder models. Similar to questions in QA methods, current prompts applied in EAE models are manual designed, based on the name of arguments or event ontology. \citet{li-etal-2021-document} make use of the conditional generation ability of BART \cite{lewis-etal-2020-bart} to generate arguments in masked position directly. While \citet{lin2022cup} incorporate curriculum learning into prompt tuning to better capture the argument level interaction. \citet{ma-etal-2022-prompt} achieve the state-of-the-art performance by adding a span selector in the model, which is a return to classic extraction methods. However, all these previous methods haven't answer the question that how prompts facilitate extraction and what kind of the bias is brought by prompts. In our work, we render a possible answer that it is the syntactic structure that the ontology-based prompts methods rely on for sound extraction. We also analyse the potential risks from ontology-based prompts in a causal view.

\subsection{Causal Inference}
Causal inference is a process of determining the independent, actual effect of a particular phenomenon \cite{pearl2009causal}, which has been applied in various of domains like psychology, politics and epidemiology for years \cite{mackinnon2007mediation,keele2015statistics}. Recently, causal inference has attracted increasingly attention from the natural language processing (NLP) community. A growing number of inspiring works have been accomplished with the assistance of causal inference. For example, \citet{cao-etal-2022-prompt} utilize structural causal model as an analysing tool to discover the potential risks hidden in probing tasks and \citet{zhang-etal-2022-de} investigate the spurious risks in generative named entity recognition (NER) model and use data augmentation to relief the problems. Other works explore the potential of counterfactual in NLP tasks, \citet{yang-etal-2021-exploring,joshi-he-2022-investigation} investigate the proper way to generate counterfactual data augmentation for a robust NLP model. Compared to them, to our best knowledge, we are the first to design debias prompt-based trainable model, with an insight on the philosophy of how manual prompts work. Our method offers a fundamental approach to eliminate the influence from the confounder in training stage.


\newpage
\clearpage
\bibliographystyle{named}
\bibliography{custom,anthology}

\end{document}